# Trajectory Planning for Autonomous Parking in Complex Environments: A Tunnel-based Optimal Control Approach *

Bai Li, Tankut Acarman, Qi Kong, and Youmin Zhang

*Abstract*—This paper proposes a fast and accurate trajectory planning algorithm for autonomous parking. Nominally, an optimal control problem should be formulated to describe this scheme, but the dimensionality of the optimal control problem is usually large, because the vehicle needs to avoid collision with every obstacle at every moment during the entire dynamic process. Although an initial guess obtained by a sample-and-search based planner facilitates the numerical optimization process, it is still far from being as fast as real-time. To address this issue, we replace all of the collision-avoidance constraints by series of within-tunnel conditions. Concretely, we develop a tunnel-based strategy such that the vehicle is restricted to move within the tunnels which naturally separate the vehicle from the obstacles. Unification, efficiency, and robustness of the proposed trajectory planning method have been verified by simulations.

## I. INTRODUCTION

### A. Background

Autonomous vehicle technologies are bringing about revolutionary changes to the urban transport [1]. As a necessary module in an autonomous driving system, trajectory planning is about generating a trajectory that is kinematically feasible for the vehicle, comfortable for the passengers, and collision-free from the detected obstacles [2]. This paper focuses on trajectory planning in the autonomous parking scenarios.

Trajectory planning algorithms for parking are more challenging than the ones for on-road driving, because (i) there is not a reference line for navigation; (ii) the vehicle kinematics should support driving backwards; and (iii) the intricate obstacles in the environment complicate the problem formulation. These factors make the prevalent on-road trajectory planners not directly applicable to the parking tasks.

### B. Related Works

Broadly speaking, the trajectory planners that are capable of handling autonomous parking schemes can be classified as sample-and-search-based, and optimization-based methods. A sample-and-search-based planner first abstracts the continuous state space as a graph, then searches in the graph for satisfactory nodes that link the starting and goal configurations. This category can be further divided by sampling the state space or the input space. Typical state-space samplers include the state lattice planner [3] and the Rapidly-exploring Random Tree (RRT) families [4]. Well-known input-space samplers include the hybrid A* algorithm [5] and Dynamic Window Approach (DWA) [6]. Optimization-based methods, on the other hand, describe the concerned trajectory planning scheme as an optimal control problem, and then numerically solve it. The numerical solution is derived by converting the optimal control problem into a nonlinear programming (NLP) problem, and then solving that NLP. NLP solvers such as sequence quadratic programming (SQP) [7,8], interior-point method (IPM) [9], convex feasible set algorithm [10], and $g^2o$ [11] have been applied to parking oriented trajectory planning problems.

Compared with the sample-and-search-based planners, formulating an optimal control problem is advantageous because (i) the continuous state space needs not discretized into primitives; and (ii) trajectories are directly planned without path velocity decomposition. But the side effect of an optimization-based planner is the heavy computational burden. Typically an optimal control problem contains large numbers of non-convex collision-avoidance constraints, which hinder the online applications. A common remedy for this limitation is utilizing a sample-and-search-based planner to quickly find a coarse path/trajectory, then implementing numerical optimization with that initial guess [11–13]. Although a quickly searched initial guess facilitates the numerical solution process, the large-scale non-convex constraints still exist in the formulated optimal control problem, which make the optimization slow *whenever* the initial guess is not close to the optimum. Therefore, in addition to maintaining the initialization quality, substantial efforts are needed to simplify the optimal control problem formulation.

The primary difficulties in the concerned optimal control problem lie in the collision-avoidance constraints [14]. Since the vehicle may not have chances to collide with every obstacle in the environment, some of the collision-avoidance constraints can be safely removed, especially when a coarse path/trajectory is given (Fig. 1(a)). Following this idea, it is natural to consider paving a tunnel that separates the vehicle from all of the surrounding obstacles. With such a tunnel at hand, one can use the within-tunnel constraints to completely replace the collision-avoidance constraints. Through this, the dimensionality and non-convexity of the optimal control problem becomes independent from the complexity of the

* Research supported by the National Key R&D Program of China under Grant 2018YFB1600804, and the Natural Sciences and Engineering Research Council of Canada.

Bai Li is with the JDX R&D Center of Automated Driving, JD Inc., Beijing, China (e-mail: libai@zju.edu.cn, libai1@jd.com).
Tankut Acarman is with the Computer Engineering Department, Galatasaray University, Istanbul, Turkey (e-mail: tacarman@gsu.edu.tr).
Qi Kong is with the JDX R&D Center of Automated Driving, JD Inc., Beijing, China (e-mail: Qi.Kong@jd.com).
Youmin Zhang is with the Department of Mechanical, Industrial and Aerospace Engineering, Concordia University, Montreal, Canada (e-mail: ymzhang@encs.concordia.ca).

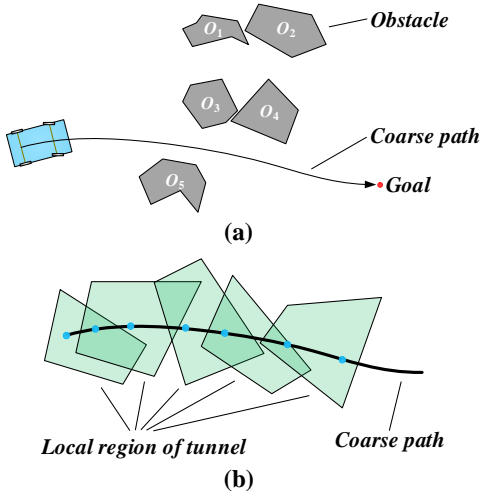

Fig. 1. Schematics on tunnel-based strategy: (a) since the vehicle has chosen one homotopic coarse path, it has slim chances to collide with $O1$ or $O2$; (b) assuming that the vehicle is a mass point, previous methods require that the vehicle stays in one of the local regions at every moment, but a ground vehicle cannot be treated as a mass point, thus previous methods are not applicable to describe the possibilities that a ground vehicle stays in more than one local regions.

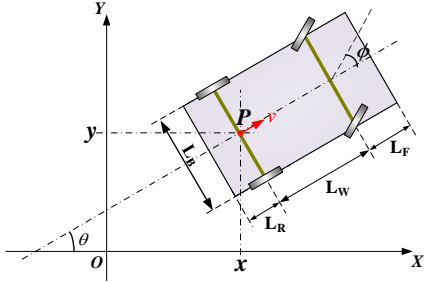

Fig. 2. Schematics on vehicle kinematics and geometrics.

environment. The tunnel-based strategy was applied in the aerial vehicle trajectory planning [15–17]. However, since the shape of a ground vehicle cannot be simply treated as a mass point, it is possible that the vehicle body covers multiple local regions of the tunnel (Fig. 1(b)). This new challenge makes those previous methodologies not applicable here.

The only reference that considered the tunnel-based strategy for autonomous parking problems, to the best of our knowledge, is [18], wherein local neighborhoods of a coarse path/trajectory is created so that the within-tunnel conditions are simply formulated as bounds imposed on the vehicle's 2D location and the orientation angle. Although being able to describe the within-tunnel conditions as the simplest type of linear constraints, that method requires extremely extensive offline efforts to create the neighborhood templates according to the vehicle's shape. In addition, that method blindly converts the angular scale to a distance scale through a pre-defined weighting parameter when deciding the size of the neighborhood, thus it suffers from the risk to lose the solution feasibility.

### C. Contributions

This paper aims to plan fast, accurate, and near-optimal trajectories for generic autonomous parking problems. To this end, an optimal control problem is formulated, wherein the collision-avoidance constraints are simplified through a tunnel construction strategy. Compared with the previous works, our tunnel-construction strategy (i) does not require any offline preparation efforts before the online usage; and (ii) addresses the issue that different parts of the vehicle may stay in different local regions of the tunnel. With our novel tunnel-based strategy, the scale of the formulated optimal control problem is completely independent from the complexity of the environment, which is actually a desirable property in online trajectory planning.

### D. Organization

In the rest of this paper, Section II briefly defines the trajectory planning oriented optimal control problem and the numerical solution principle. Section III introduces the way to describe the collision-avoidance restrictions as within-tunnel constraints through the tunnel-based strategy. Simulation results and discussions are provided in Section IV. Finally, the conclusions are drawn in Section V.

## II. OPTIMAL CONTROL PROBLEM DEFINITION AND SOLUTION

This section provides the overall optimal control problem formulation for describing the autonomous parking trajectory planning scheme, and introduces how to solve the problem numerically.

### A. Optimal Control Problem Formulation

The parking trajectory planning oriented optimal control problem consists of a cost function, vehicle kinematics, boundary conditions and the within-tunnel constraints.

Since a vehicle usually runs at low speeds during the parking process, the bicycle model is sufficient to describe the vehicle kinematics [9]:

$$\frac{\mathrm{d}}{\mathrm{d}t}\begin{bmatrix} x(t) \\ y(t) \\ v(t) \\ \phi(t) \\ \theta(t) \end{bmatrix} = \begin{bmatrix} v(t)\cdot\cos\theta(t) \\ v(t)\cdot\sin\theta(t) \\ a(t) \\ \omega(t) \\ v(t)\cdot\tan\phi(t)/\mathrm{L_W} \end{bmatrix}, \; t \in [0, t_\mathrm{f}]. \quad (1)$$

Herein, $t_\mathrm{f}$ denotes the parking process duration (not fixed), $(x, y)$ represents the mid-point of rear wheel axis (see point $P$ in Fig. 2), $v$ represents the velocity of $P$, $a$ represents the corresponding acceleration profile, $\phi$ refers to the steering angle, $\omega$ represents the corresponding angular velocity, $\theta$ refers to the orientation angle, and $\mathrm{L_W}$ denotes the wheelbase. In addition to $\mathrm{L_W}$, other geometric parameters such as $\mathrm{L_F}$ (front overhang length), $\mathrm{L_R}$ (rear overhang length), and $\mathrm{L_B}$ (width) are also depicted in Fig. 2. Boundaries are imposed on some of the aforementioned profiles to describe the mechanical/physical principles of the movement:

$$\begin{aligned} |a(t)| &\le \mathrm{a_{max}} \\ |v(t)| &\le \mathrm{v_{max}} \\ |\omega(t)| &\le \Omega_{\max} \\ |\phi(t)| &\le \Phi_{\max} \end{aligned}, \; t \in [0, t_\mathrm{f}]. \quad (2)$$

Boundary conditions specify the vehicle's configurations at the initial and terminal moments:

$$\begin{bmatrix} x(0) \\ y(0) \\ \theta(0) \\ v(0) \\ \phi(0) \\ a(0) \\ \omega(0) \end{bmatrix} = \begin{bmatrix} x_{init} \\ y_{init} \\ \theta_{init} \\ v_{init} \\ \phi_{init} \\ a_{init} \\ \omega_{init} \end{bmatrix}, \begin{bmatrix} x(t_f) \\ y(t_f) \\ \theta(t_f) \\ v(t_f) \\ \phi(t_f) \\ a(t_f) \\ \omega(t_f) \end{bmatrix} = \begin{bmatrix} x_{final} \\ y_{final} \\ \theta_{final} \\ v_{final} \\ \phi_{final} \\ a_{final} \\ \omega_{final} \end{bmatrix}, \qquad (3)$$

wherein $x_{init}$, $y_{init}$, $\theta_{init}$, $v_{init}$, $\phi_{init}$, $a_{init}$, $\omega_{init}$, $x_{final}$, $y_{final}$, $\theta_{final}$, $v_{final}$, $\phi_{final}$, $a_{final}$, and $\omega_{final}$ are parameters which determine the starting and terminal configurations.

The within-tunnel constraints are utilized to avoid the collision risks with the obstacles. The details are introduced in the next section.

Regarding the cost function of the optimal control problem, both efficiency and comfort are considered. Concretely, efficiency is reflected by the expectation to accomplish the parking movement subject to minimum time, and comfort is reflected by the expectation to spend minimum energy to change $v$ and $\phi$:

$$J_{cost} = t_f + w_1 \cdot \int_{t=0}^{t_f} a^2(t) \cdot dt + w_2 \cdot \int_{t=0}^{t_f} \omega^2(t) \cdot dt, \qquad (4)$$

wherein $w_1, w_2 \geq 0$ are weighting parameters. As a summary, the autonomous parking trajectory planning task is described as the following optimal control problem:

$$\begin{aligned} & \text{Minimize (4),} \\ & \text{s.t. Kinematic constraints (1), (2);} \\ & \quad \text{Within-tunnel constraints;} \\ & \quad \text{Boundary conditions (3).} \end{aligned} \qquad (5)$$

### B. Numerical Solution to Optimal Control Problem

The unknowns in (5) include $x(t)$, $y(t)$, $\theta(t)$, $v(t)$, $a(t)$, $\omega(t)$, $\phi(t)$, and $t_f$. If $\omega(t)$, $a(t)$, and $t_f$ are determined, the rest profiles are uniquely fixed. Instead of optimizing only $\omega(t)$, $a(t)$, and $t_f$ like a shooting method, this work adopts the collocation method which regards all of the state and control profiles as decision variables. A collocation method typically achieves high-level solution stability in contrast to a shooting method [19].

The first-order explicit Runge-Kutta method is applied to discretize the 7 profiles $x(t)$, $y(t)$, $\theta(t)$, $v(t)$, $a(t)$, $\omega(t)$, $\phi(t)$, as well as the cost function/constraints. Through this discretization, an NLP is built. SQP is chosen as the NLP solver because it is more warm-starting friendly than a barrier-function method (such as IPM). Finally, the output of SQP is the planned trajectory for parking.

## III. WITHIN-TUNNEL CONSTRAINT FORMULATION

This section formulates the within-tunnel constraints in the optimal control problem (5).

### A. Step 0. Dilating Obstacles and Shrinking Vehicle Body

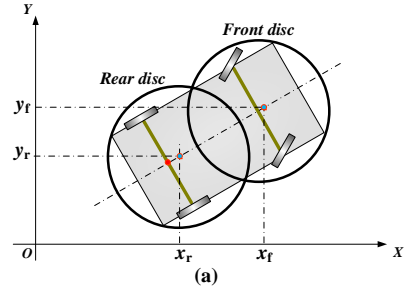

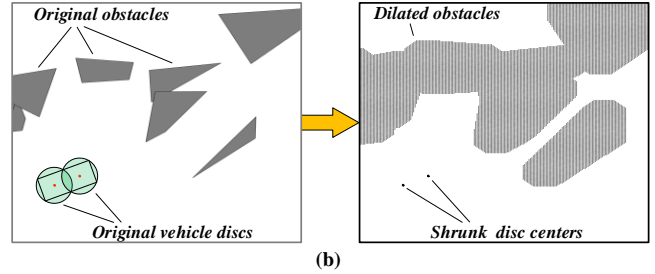

Fig. 3. Schematics on vehicle shape shrink and environmental map dilatation: (a) presenting vehicle shape with two discs; (b) simultaneously converting the vehicle shape and environmental obstacles into new forms.

As a common practice, discs can be used to cover the rectangular vehicle body [20]. As depicted in Fig. 3(a), this work adopts 2 discs to cover the vehicle body. Denoted as $P_f = (x_f, y_f)$ and $P_r = (x_r, y_r)$, the centers of the two discs are quartile points along the vehicle's longitudinal axis, i.e.,

$$\begin{aligned} x_f(t) &= x(t) + \frac{1}{4}(3L_W + 3L_F - L_R) \cdot \cos\theta(t), \\ y_f(t) &= y(t) + \frac{1}{4}(3L_W + 3L_F - L_R) \cdot \sin\theta(t), \\ x_r(t) &= x(t) + \frac{1}{4}(L_W + L_F - 3L_R) \cdot \cos\theta(t), \\ y_r(t) &= y(t) + \frac{1}{4}(L_W + L_F - 3L_R) \cdot \sin\theta(t). \end{aligned} \qquad (6a)$$

The radius of each disc, denoted as $R_C$, is determined via

$$R_C = \frac{1}{2}\sqrt{(\frac{L_R + L_W + L_F}{2})^2 + (L_B)^2}. \qquad (6b)$$

Nominally the collision-avoidance constraints require that each disc keeps clear of the obstacles, which is identical to the restriction that each disc center should keep at least a distance of $R_C$ from the obstacles. This means we can make an equivalent conversion to simultaneously shrink the two discs as their centers (i.e. $P_f$ and $P_r$) and dilate the obstacles by $R_C$ (Fig. 3(b)). Although this equivalent conversion alone does not make any inherent change to the planning task, it contributes to the formulation of our within-tunnel constraints, which will be introduced in the next few subsections.

As a summary of this subsection, the original trajectory planning scheme is converted into a new form: the vehicle body is shrunk to two mass points, and the environmental obstacles are dilated by $R_C$. In the rest of the paper, we refer to the environmental map with dilated obstacle as *dilated map*.

## B. Step 1. Generating Reference Trajectories

This subsection aims to generate the reference trajectories of $P_f$ and $P_r$ for future usages. To this end, we first adopt the hybrid A* algorithm to find a path for point $P=(x,y)$, then attach a time course along that path to form a trajectory $Traj_P$, and then uniquely determine the trajectories of $P_f$ and $P_r$ according to $Traj_P$. Besides that, $Traj_P$ is also used as the initial guess in the NLP solution process.

The reason of choosing the hybrid A* algorithm is it respects the kinematic model and supports backward maneuvers. In this preliminary work, we assume that there are no predictable or tractable moving obstacles in the parking scenario, then the procedure to attach a time course along the path derived by hybrid A* becomes a one-dimensional minimum-time optimal control problem, which can be analytically solved via Pontryagin's maximum principle. Once $Traj_P$ is determined, the trajectories of $P_f$ and $P_r$ are uniquely determined according to (6a). Let us denote the time domain of $Traj_P$ as $t \in \left[0, \tilde{t}_f\right]$, and the trajectories of $P_f$ and $P_r$ as $Traj_{P_f}$ and $Traj_{P_r}$, respectively.

Nominally, $P_f$ and $P_r$ should avoid collisions with any of the dilated obstacles in the dilated map. Now with the reference trajectories $Traj_{P_f}$ and $Traj_{P_r}$ at hand, we no longer need to formulate the collision-avoidance constraints. Instead, we only need to require that $P_f$ and $P_r$ stay in two tunnels that are homotopic with $Traj_{P_f}$ and $Traj_{P_r}$, respectively. Since the tunnels are non-convex, we use local convex boxes to cover each tunnel. Those local boxes are referred to as *representative boxes* in the rest of this paper.

## C. Step 2. Specifying Representative Boxes

In this subsection, two series of representative boxes are generated to cover $Traj_{P_f}$ and $Traj_{P_r}$, respectively. Let us focus on the box generation scheme for $Traj_{P_f}$ at first.

To begin with, we sample $(N_R+1)$ waypoints along $Traj_{P_f}$ evenly in the time horizon. Concretely, the waypoints $(x_f(t), y_f(t))$ at $t = \dfrac{\tilde{t}_f}{N_R} \cdot i$ $(i=0,1,...,N_R)$ are extracted from $Traj_{P_f}$. These waypoints are referred to as *representative nodes* (see the example illustrated in Fig. 4). Once the $(N_R+1)$ representative nodes are available, the next step is to specify $(N_R+1)$ representative boxes in association with the representative nodes. The principle to specify the *k*th representative box is introduced as follows.

Firstly, we use the finite difference method to specify the orientation angle $\theta_f(t_k)$ at the currently concerned representative node $(x_f(t_k), y_f(t_k))$ along $Traj_{P_f}$, where $t_k = \dfrac{\tilde{t}_f}{N_R} \cdot k$. Secondly, we define four direction, namely $\theta_f(t_k)$,

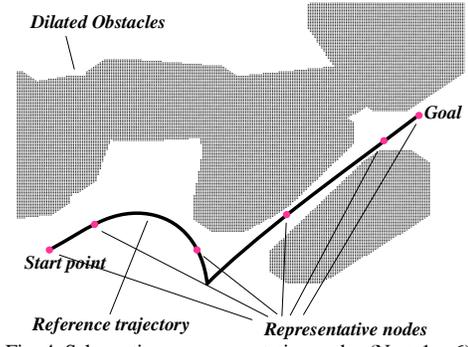
Fig. 4. Schematics on representative nodes ($N_R + 1 = 6$).

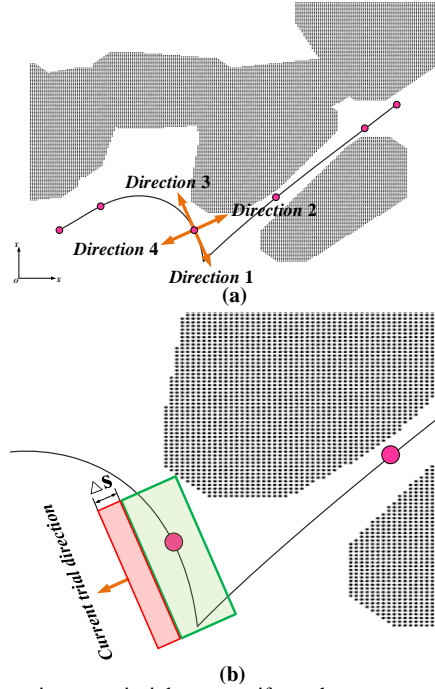
Fig. 5. Schematics on principle to specify each representative box: (a) expansion directions; (b) expansion trial in one direction.

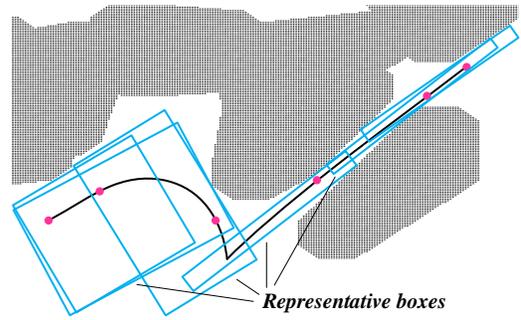
Fig. 6. Schematics on tunnel formed by representative boxes. Ideally these representative boxes should fully cover the reference trajectory.

$\theta_f(t_k) + \dfrac{\pi}{2}$, $\theta_f(t_k) + \pi$, and $\theta_f(t_k) + \dfrac{3\pi}{2}$ (see Fig. 5(a)). We regard the representative node as a zero-width-zero-length rectangle (i.e. a core), and incrementally expand the core in each of the four directions by a constant step $\Delta s$ each time. Whenever an expansion trial in one direction is made, we check if that trial causes collisions with the obstacles in the dilated map. If no collision occurs, then that trial is approved; otherwise it is rejected and future expansion trials in that direction are not considered any longer. Let us take Fig. 5(b) as an example, suppose the green box denotes the currently approved region, and the expansion trial in the currently concerned direction renders the red box. Since the red box does not overlap with the obstacles in the dilated map, the red box is approved, which means the approved region now consists of both the green and red boxes. Besides that, we define a maximum expansion length parameter $L_{limit}$ to avoid excessive expansions in each direction. The principle to specify the geometric size of the *k*th representative box is summarized in the following pseudo code.

**Algorithm 1.** Representative box specification approach.

*Input*: Dilated map, $[x_f(t_k), y_f(t_k), \theta_f(t_k)]$, and parameter $L_{limit}$ ;

*Output*: Vertex locations of representative box k.
1. Define four expansion directions according to $\theta_f(t_k)$ ;
2. Initialize an index set $\Omega = \{1,2,3,4\}$ ;
3. Initialize a vector as **length** $=[0,0,0,0]$ ;
4. Initialize approved region $\Upsilon$ as the mass point $(x_f(t_k), y_f(t_k))$ ;
5. **While** $\Omega \neq \emptyset$, **do**
6.    **For each** $i \in \Omega$, **do**
7.       Expand $\Upsilon$ in direction $i$ by $\Delta s$ to form a trial box region $\Upsilon_{trial}$ ;
8.       **If** $\Upsilon_{trial}$ does not overlap with obstacles of dilated map, **then**
9.          Update $\mathbf{length}_i \leftarrow \mathbf{length}_i + \Delta s$ ;
10.         Merge $\Upsilon_{trial}$ into $\Upsilon$ ;
11.         **If** $\mathbf{length}_i > L_{limit}$, then
12.            Remove $i$ from $\Omega$ ;
13.         **End if**
14.       **Else**
15.          Remove $i$ from $\Omega$ ;
16.       **End if**
17.    **End for**
18. **End while**
19. Specify four vertexes' locations according to $\Upsilon$ ;
20. Output vertex locations;
21. Return.

Through specifying all of the representative boxes in the same way, we can pave a tunnel for point $P_f$, which consists of $(N_R + 1)$ representative boxes (Fig. 6). The tunnel for $P_r$ can be paved in the same way. To avoid confusion, the tunnel for $P_f$ is denoted as the *front tunnel*, while the one for $P_r$ is denoted as the *rear tunnel*.

### D. Step 3. Formulating Within-Tunnel Constraints

Based on the preparations in the past few subsections, this subsection formally builds the within-tunnel constraints. Concretely, we require that

$P_f(t)$ stays in *k*th representative box from front tunnel

$$\text{when } t \in \left[\frac{t_f}{N_R} \cdot k, \frac{t_f}{N_R} \cdot (k+1)\right], k = 0,...,N_R - 1; \quad (7a)$$

$P_f(t_f)$ stays in $N_R$th representative box from front tunnel.

Similarly, we impose the following constraints for $P_r$:

$P_r(t)$ stays in *k*th representative box from rear tunnel

$$\text{when } t \in \left[\frac{t_f}{N_R} \cdot k, \frac{t_f}{N_R} \cdot (k+1)\right], k = 0,...,N_R - 1; \quad (7b)$$

$P_r(t_f)$ stays in $N_R$th representative box from rear tunnel.

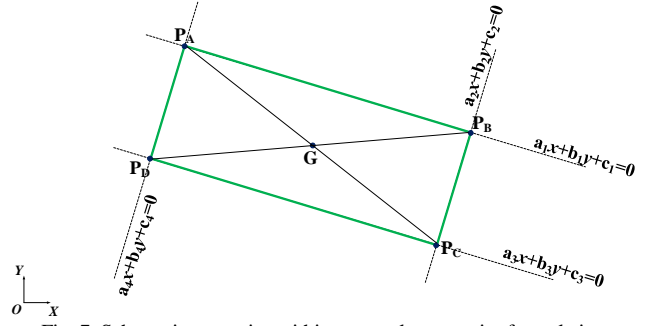

Fig. 7. Schematics on point-within-rectangle constraint formulation.

Next, we will briefly introduce how to describe such restrictions as algebraic inequalities.

Restricting a point S to stay in an irregularly placed rectangle is identical to the condition that the point S simultaneously stays on some one side of the rectangle's each edge. Taking Fig. 7 as an example, we denote the four vertexes of the rectangle as $P_A$, $P_B$, $P_C$, and $P_D$, and denote point S as $(x_S, y_S)$. Now the within-rectangle constraint is described as the restriction that point S should locate within the region surrounded by straight lines $P_AP_B$, $P_BP_C$, $P_CP_D$, and $P_DP_A$. Each straight line can be presented in the form of an equality. For example, line $P_AP_B$ is described as $a_1 \cdot x + b_1 \cdot y + c_1 = 0$ with

$$\begin{aligned} a_1 &= y_2 - y_1, \\ b_1 &= x_1 - x_2, \\ c_1 &= x_2 \cdot y_1 - x_1 \cdot y_2, \\ P_A &= (x_1, y_1), \ P_B = (x_2, y_2). \end{aligned} \quad (8)$$

Requiring that S stays on one side of line $P_AP_B$ can be described as $a_1 \cdot x_s + b_1 \cdot y_s + c_1 > 0$ or $a_1 \cdot x_s + b_1 \cdot y_s + c_1 < 0$. Regarding how to choose one between the two inequalities, we notice that point S and the rectangle's geometric center (denoted as $G = (x_{center}, y_{center})$ in Fig. 7) stay on the same side of line $P_AP_B$. Therefore, if $a_1 \cdot x_{center} + b_1 \cdot y_{center} + c_1 > 0$, then we choose $a_1 \cdot x_s + b_1 \cdot y_s + c_1 > 0$, and vice versa. The constraints in association with the rest three straight lines are specified in the same way. As a conclusion, four linear inequalities constitute the point-within-rectangle constraint.

Regarding the trajectory planning scheme we concern, $8 \cdot (N_R + 1)$ types of simple inequalities totally constitute the within-tunnel constraints. Apparently, the scale of the constraints is irrelevant to the number of obstacles in the environment. In contrast with the nominal collision-avoidance constraints (e.g. in [21]) which are non-differentiable and non-convex, our within-tunnel constraints are nearly linear, thereby being easy to be handled by the NLP solver.

Before the end of this section, we would like to emphasize that although we only require the two points (namely $P_f$ and $P_r$) to travel in the tunnels, the tunnels are generated according to the dilated map rather than the original map, thus it is safely guaranteed that the whole vehicle body keeps clear of the obstacles *provided that* $P_f$ *and* $P_r$ *stay in their own tunnels*.

## IV. SIMULATION RESULTS AND DISCUSSIONS

Simulations were performed in C++ and executed on an i7-7700 CPU with 8 GB RAM that runs at 3.60 × 2 GHz. SNOPT, a commercial software package of SQP was utilized in AMPL with default options. MATLAB 2019a was used to demonstrate the simulation results. Basic parametric settings are listed in Table I. A video with the primary simulation results is provided at https://youtu.be/brQo9lPw9cw.

TABLE I. PARAMETRIC SETTINGS REGARDING MODEL AND APPROACH.

| Parameter | Description | Setting |
|---|---|---|
| $L_F$ | Front hang length of vehicle. | 0.96 m |
| $L_W$ | Wheelbase of vehicle. | 2.80 m |
| $L_R$ | Rear hang length of vehicle. | 0.929 m |
| $L_B$ | Width of vehicle. | 1.942 m |
| $a_{max}$ | Upper bound of $|a(t)|$. | 4.0 m/s$^2$ |
| $v_{max}$ | Upper bound of $|v(t)|$. | 3.0 m/s |
| $\Phi_{max}$ | Upper bound of $|\phi(t)|$. | 0.70 rad |
| $\Omega_{max}$ | Upper bound of $|\omega(t)|$. | 0.5 rad/s |
| $w_1, w_2$ | Weights in cost function (4). | 0.1, 0.01 |
| $N_R + 1$ | Number of representative boxes in each tunnel. | 61 |
| $\Delta s$ | Unit step length in Algorithm 1. | 0.1 m |
| $L_{limit}$ | Maximum step length in Algorithm 1. | 8.0 m |
| $N_{fe}$ | Number of finite elements in Runge-Kutta method. | 60 |

### A. On the Efficacy of Trajectory Planner

The first round of simulations focuses on the efficacy of the planned trajectories. The optimized trajectories of three parking cases are depicted in Figs. 8–10, respectively. Cases 1 and 2 represent the scenarios with irregularly parked cars near our ego-vehicle, while Case 3 represents a cluttered environment. According to the footprints in Figs. 8–10, the ego-vehicle manages to avoid collisions with the obstacles, which show the efficacy of the proposed planner. Particularly, the optimized profiles $v(t)$ and $\phi(t)$ in Case 1 are shown in Fig. 11, which reflect satisfactions to the vehicle kinematic restrictions (1) and (2).

### B. On the Efficacy of Tunnel-based Strategy

The second round of simulations investigates the efficacy of the proposed tunnel-based strategy. Let us take Case 2 as an example. Fig. 12 plots the coarse path derived by the hybrid A* algorithm, the optimized trajectories with the tunnel-based strategy under various settings of $N_R$, and the local optimum derived by the numerical optimal control approach [21] with complete collision-avoidance constraints. Compared with the coarse path obtained by the hybrid A* algorithm, the optimized trajectories are smoother. In contrast with the local optimum derived by [21], the trajectories obtained with the tunnel-based strategy are not optimal, and there is not a trend that the solution converges to the local optimum as $N_R$ grows. In order to have a straightforward impression of the paved tunnels, Fig. 13 illustrates the tunnels with $N_R = 40$ and $N_R = 200$. In that figure, it is obvious that part of the drivable area are not covered by the representative boxes. This phenomenon may be regarded as a limitation of this work, and using other types of representative polygons would improve the situation but Algorithm 1 becomes more complicated then.

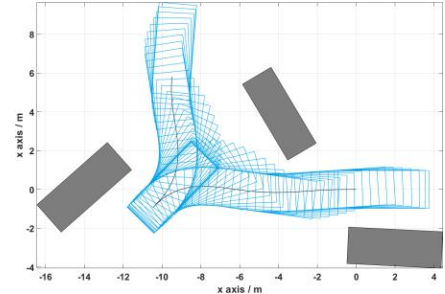

Fig. 8. Optimized parking trajectory and footprints in Case 1 ($t_f = 13.9555$ s).

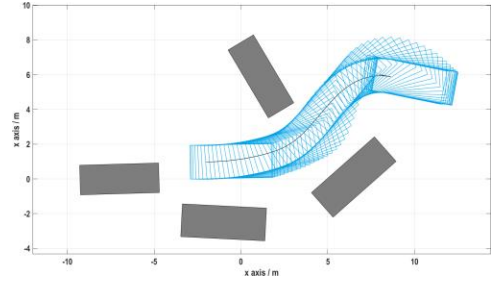

Fig. 9. Optimized parking trajectory and footprints in Case 2 ($t_f = 10.3821$ s).

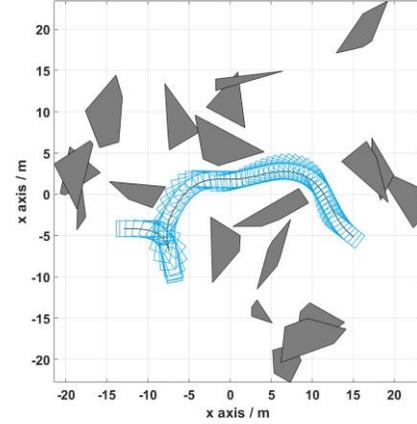

Fig. 10. Optimized parking trajectory and footprints in Case 3 ($t_f = 21.6495$ s).

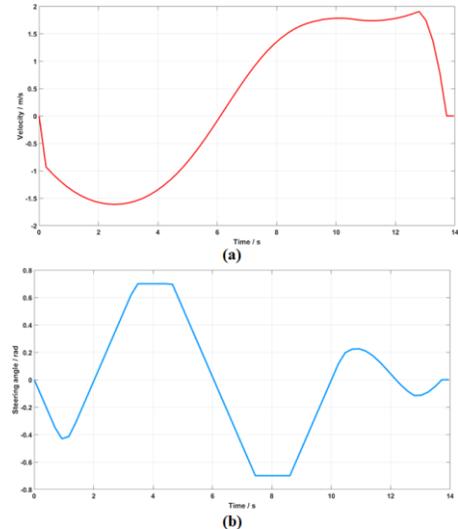

Fig. 11. Optimized profiles in Case 1: (a) $v(t)$, and (b) $\phi(t)$.

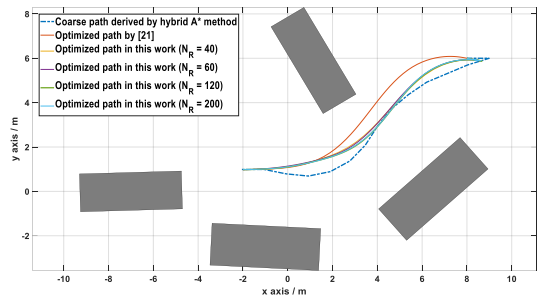

Fig. 12. Optimized trajectories with various settings of $N_R$ (Case 2).

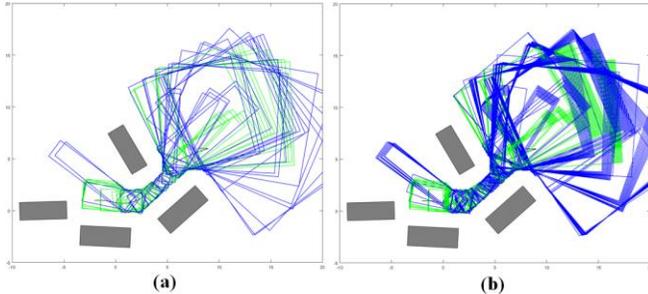

Fig. 13. Paved tunnels with various settings of $N_R$ in Case 2. Node that the blue boxes denote the representative boxes for $P_f$, while green for $P_r$.

Through comparing among the optimized trajectories with our tunnel-based strategy, we notice that the changes in $N_R$ do not alter much in the solutions, which reflects the robustness of the proposed tunnel-based strategy.

## V. CONCLUSIONS

This paper has proposed a fast trajectory planner for generic autonomous parking schemes. Compared with the prevalent planners that formulate the large-scale and complicated collision-avoidance constraints, we consider paving tunnels which naturally separate the vehicle body from the obstacles. The proposed tunnel-based strategy makes the scale of the optimal control problem insensitive to the complexity of the environment.

As our future work, (i) the parking cases with moving obstacles will be considered; (ii) other types of convex polygons rather than rectangles may be adopted for covering the tunnels; (iii) the optimized profile in Fig. 11(a) indicates a need to impose bounds on jerk. We will also try the mixed integer mathematical programming formulations introduced in [15]–[17] for potential chances of promoting the solution optimality.